\documentclass{article}
\usepackage[T1]{fontenc}
\usepackage[utf8]{inputenc}
\usepackage{array}
\usepackage{float}
\usepackage{algorithm2e}
\usepackage{amsmath}
\usepackage{graphicx}

\makeatletter

\providecommand{\tabularnewline}{\\}

\usepackage{url}
\usepackage{booktabs}
\usepackage{amsfonts}\usepackage{amsthm}
\usepackage{nicefrac}
\usepackage{microtype}
\usepackage{algorithmic}
\usepackage{float}

\author{
  Tal Kachman, Michal Moshkovitz, Michal Rosen-Zvi
    \\
  IBM Research \\}

\date{}

\title{Novel Uncertainty Framework for Deep Learning Ensembles}

\renewcommand\[{\begin{equation}}
\renewcommand\]{\end{equation}}

\@ifundefined{showcaptionsetup}{}{%
 \PassOptionsToPackage{caption=false}{subfig}}
\usepackage{subfig}
\makeatother

\begin{document}
\maketitle 
\begin{abstract}
Deep neural networks have become the default choice for many of the
machine learning tasks such as classification and regression. Dropout,
a method commonly used to improve the convergence of deep neural networks,
generates an ensemble of thinned networks with extensive weight sharing.
Recent studies that dropout can be viewed as an approximate variational
inference in Gaussian processes, and used as a practical tool to obtain
uncertainty estimates of the network. We propose a novel statistical
mechanics based framework to dropout and use this framework to propose
a new generic algorithm that focuses on estimates of the variance
of the loss as measured by the ensemble of thinned networks. Our approach
can be applied to a wide range of deep neural network architectures
and machine learning tasks. In classification, this algorithm allows
the generation of a don't-know answer to be generated, which can increase
the reliability of the classifier. Empirically we demonstrate state-of-the-art
AUC results on publicly available benchmarks.
\end{abstract}

\section{Introduction}

Deep learning (DL) algorithms have successfully solved real-world
classification problems from a variety of fields, including recognizing
handwritten digits and identifying the presence of key diagnostic
features in medical images \cite{lisboa2006use, lecun1990handwritten}.
A typical classification challenge for a DL algorithm consists of
training the algorithm on an example data set, then using a separate
set of test data to evaluate its performance. The aim is to provide
answers that are as accurate as possible, as measured by the true
positive rate (TPR) and the true negative rate (TNR). Many DL classifiers,
particularly those using a softmax function in the very last layer,
yield a continuous score, $h$; A step function is used to map this
continuous score to each of the possible categories that are being
classified. TPR and TNR scores are then generated for each separate
variable that is being predicted by setting a threshold parameter
that is applied when mapping $h$ to the decision. Values above this
threshold are mapped to positive predictions, while values below it
are mapped to negative predictions. The ROC curve is then generated
from these pairs of TPR/TPN scores. The performance of binary classifiers,
performance is often evaluated by calculating the area under the ROC
curve (AUC) \cite[{hanley1982meaning}]; the higher the AUC, the better
the performance of the algorithm. 

Many studies show that the AUC achieved by DL algorithms is higher
than most, if not all, of the alternative classifiers. Although they
achieve high scores for metrics such as AUC, DL algorithms are notorious
for being ``black box'' models, as it is difficult to obtain insight
into how the algorithm arrived at its conclusion. This makes them
less reliable, particularly for applications where data-based decisions
are a firm requirement. One way to mitigate this problem when applying
DL algorithms to these data-driven applications is to provide a measure
of the classification uncertainty, or the confidence one has in the
classification prediction, along with the prediction of the outcome.
This can be accomplished by training a series of similar models that
differ in their initialization parameters, and then calculating the
variance between the probabilities for each possible outcome across
the differently initialized networks. The ground truth of the uncertainty,
then, is the variance between these similar models. Reporting a confidence
level along side the prediction helps the model's end user to interpret
and build trust in the model's performance. For Example, this is a
critical requirement for algorithms deployed in a medical setting
\cite{bellazzi2008predictive}, where a moderately positive prediction
with high uncertainty has very different prognosis and treatment implications
than does the same prediction when the uncertainty is low. We show
an example of both low variance and high variance predictions, calculated
using the ground truth as defined , in Figure~\ref{fig:higher_moments}. 

When assessing uncertainty in DL algorithms such classifiers are frequently
permitted to return a ``don't know'' answer for very low confidence
predictions in the test data. This allows the algorithm to be judged
only on the responses in the ``do know'' portion of the data set.
Consequently, the algorithm generates overall higher quality predictions,
while leaving humans to interpret samples for which it would generate
poor quality predictions, similar to a triage system 

The continuous variable $h$ that is output by the last layer of a
DL with a softmax function provides the likelihood of an outcome.
However, this likelihood estimation should not be mistaken for a confidence
level in the prediction (see \cite{gal2016uncertainty}). Instead,
recent studies have combined the DL regularization technique of dropout
with Bayesian modeling to derive uncertainty estimates in DL classifiers
\cite{gal2016uncertainty, gal2016dropout}. Bayesian approaches provides
a natural framework for estimating the uncertainty of the prediction.
Furthermore, dropout is a regularization technique that uses an ensemble
of models to create high-performance classifiers \cite{srivastava2014dropout, lecun2015deep}.
The interpretation of dropout through a Bayesian lens enables derivation
of uncertainty estimates in DL classifiers. Within this framework
the classification score, $h$, describes the probability of predicting
the correct class. The ROC curve for the model is measured on an averaged
value of $h$, rather than a single value, and the uncertainty is
estimated based on the variance of $h$. This uncertainty measure
has been shown to correlate with the performance of the classifier.
Using the averaged $h$ to predict the outcome was sufficient to increase
classification accuracy in some cases \cite{kendall2015bayesian}.

In this paper, we present a novel method for estimating the uncertainty
of a DL classifier. We propose a framework that assigns probability
distributions to thinned networks,(i.e., neural networks with a subset
of neurons removed), based on the performance of their cross-entropy
loss function across the test data set. The main contributions of
this work are threefold. First, we introduce a statistical-mechanics
framework that assigns probability distributions over the ensemble
of thinned networks of the dropout. This framework has a flexible
variable, $\beta$, which represents the inverse temperature, and
a statistical fluctuation scale. When set to zero, it results in a
uniform distribution assumption over the thinned networks and the
framework collapses to a Gaussian process. In contrast, a finite inverse
temperature $\beta$ results in a non-uniform distribution, and the
framework enables interpretation and reasoning regarding uncertainty.
Second, we present a new algorithm, called Loss Variance Monte Carlo
Estimate (LoVME), which is based on estimations of the loss variance
in the case of a finite $\beta$ through Monte Carlo sampling. Finally,
we illustrate the benefits of deriving uncertainty through the LoVME
algorithm in scenarios where the classifier can yield a don’t-know
answer. We use the MNIST \cite{lecun1998mnist} and CIFAR \cite{krizhevsky2009learning}
data sets to show the performance of our algorithm, and compare our
results to state-of- the-art algorithms for uncertainty in DL. 

The rest of the paper is organized as follows: First, we introduce
related work. We focus on two related bodies of knowledge: recent
proposed methods to derive uncertainty measures for test data using
the Bayesian interpretation of dropout, and the existing statistical-mechanics
frameworks for analyzing distributions over an ensemble of models.
Next, we introduce a new methodology for interpreting the ensemble
of Neural Networks (NNs) generated by dropout. We present a derivation
of a new algorithm, LoVME, which uses Monte Carlo sampling to estimate
the loss variance; through this loss variance, it provides estimates
of the predicted variable $h$ and the uncertainty of the prediction.
We evaluate the performance of the algorithm on multiple data sets
and compare it to state-of-the-art algorithms. We conclude with a
discussion of the advantages of the proposed framework and algorithm,
as well as potential extensions and open questions. 

We conclude with a discussion of the advantages of the proposed framework
and algorithm, as well as potential extensions and open questions. 

\section{Related Work}

\subsection{Dropout as a Bayesian average}

DL algorithms are based on large NNs with non-linear layers. Each
layer contains units that are composed of input variables, output
variables, and the weight vectors that connect them. While deep learning
algorithms perform exceptionally well as classifiers when trained
on large data sets, they are also known to suffer from overfitting.
Different approaches have been proposed to address this drawback,
including dropout, which is among the most successful. Dropout can
be viewed as a mechanism for learning an ensemble of closely related,
partly overlapped (sometimes called ‘thinned’) neural networks. During
training, a single fixed architecture is perturbed through a random
process of pruning or thinning; this results in an ensemble of closely
related NNs. A simple version of dropout is based on a fixed random
variable, $p$. Each of the units (hidden and visible) in a NN has
probability $p$ of being retained during a given training iteration,
and probability $1-p$ of being dropped out of the network. For units
that are dropped from the NN, incoming and outgoing connections are
also removed. The result of this random pruning process is an ensemble
of thinned NNs, where if the fixed original architecture had $N$
units, the training updates generate NNs with about $pN$ units. This
can create as many as $2^{N}$ different NNs, all with a common base
architecture. Predictions on test data should ideally use the trained
ensemble of thinned NNs to find an average prediction. However, such
a prediction method requires using an exponentially large collection
of NNs, and is therefore impractical. Instead, one common method for
employing dropout is to use the full learned NN, without any pruning
but with down-scaled weights. The original trained weights of all
units are multiplied by the factor $1/p$, and the original fixed
architecture with the scaled weights is used for inference of the
test examples.

\subsection{Bayesian uncertainty in deep learning}

Recent breakthrough studies \cite{gal2016uncertainty, gal2016dropout}
have added uncertainty estimation to DL algorithms in a way that is
both empirically and mathematically grounded. The uncertainty evaluation
uses \emph{dropout inference}, where means that the ensemble of thinned
NNs is used to find the variance of the prediction. The method has
been proven successful in a variety of domains, including reinforcement
learning \cite{gal2016dropout} and computer vision \cite{kendall2015bayesian, kendall2017uncertainties}.From
a theoretical perspective, this method uses the Bayesian inference
framework to evaluate the uncertainty. In this framework, a prior
over the model parameters and a likelihood are defined. Using Bayes'
theorem, these two distributions yield a posterior over the model
parameters. Marginalizing the posterior leads to the posterior predictive
distribution, which encapsulates the uncertainty. Unfortunately, marginalization
is computationally intractable. To overcome this difficulty, the approximation
inference paradigm \cite{fox2012tutorial} suggests replacing the
true posterior by an approximate one that is restricted to belong
to a simpler form of distribution family. The dropout inference method
uses the approximation inference paradigm for NNs. They define the
prior as a product of multivariate normal distributions for each layer,
and the likelihood as a softmax for multiclass classification. The
simpler form of distribution family is a mixture of two Gaussians
with small variances, and the mean of one of the Gaussians is fixed
at zero.

To evaluate the posterior predictive distribution, Kendall et al.
\cite{kendall2015bayesian, kendall2017uncertainties} used Monte
Carlo integration. This integration samples a few thinned NNs and
calculates their variance. This variance is used as an approximation
for the uncertainty. It is important to note that in calculating the
variance, each sampled thinned NN is of equal weight. In this paper,
we claim that this uniform averaging is only a crude estimation. Inspired
by ideas from statistical mechanics, we suggest taking into account
the (estimated) loss of each thinned NN. We demonstrate that the loss
for different thinned NNs can differ by an order of magnitude, thus
using an appropriate weighting for each thinned NN is extremely valuable 

\subsection{Statistical mechanics approach}

Work on random networks in the context of Bayesian NNs has a long
history \cite{mackay1996near,cho2009kernel}. In the Bayesian approach,
one can consider the different weights in the network as drawn from
a certain prior and trained towards a posterior. This suggests that
one can in fact look at the different thinned networks created by
dropout as an ensemble. The approach of viewing ensembles in order
to approximate the behavior and states of individual entities is a
fundamental approach in the field of statistical mechanics. Applying
techniques and models commonly used in statistical mechanics to problems
in deep learning has seen a recent resurgence. For example, Choromanska
et al. \cite{choromanska2015loss} showed that random rectified linear
NNs could, with approximation, be mapped onto spin glasses. Baity
\cite{baity2018comparing} explored the learning dynamics and effects
of training deep networks as glassy dynamics, and Schoenholz et al.
\cite{schoenholz2017correspondence} construct a full statistical
field theory for deep NNs. Consequently, drawing on methods and findings
from statistical mechanics can be a powerful tool for improving deep
learning methods.

Motived by this set of recent results, and the intrinsic link between
statistical mechanics and statistical properties of thinned NN, we
formulate a new approach to infer and obtain uncertainty measures
for different models. 

\section{Statistical Mechanics Framework for Deep Learning Ensemble}

\subsection{Methodology }

In this section, we derive an explicit connection between random NNs
and statistical mechanics. This link between the two methodologies
will provide insight into the different moments and cumulants of the
the NN ensemble. One can think of the uncertainty as the equivalent
of the statistical properties of the ensemble in the test phase. 

To describe the stochastic ensemble of networks, we must design a
measure for the probability space and define the fluctuation size
with its given statistical properties, (i.e., its observable quantity).
Following the well-known methodology of Jaynes \cite{jaynes1957information},
we can write a maximum entropy argument that allows us to obtain the
maximally unbiased probability distribution. Motivated by dropout,
we define an ensemble that is defined by the full NN and its associated
thinned NNs. Our observables will be the average loss and for a changing
number of neurons average number of units. 

For an ensemble of thinned networks, we define each thinned network
$i$, is defined by its loss $\mathcal{L}_{i}$ and the number of
neurons that it contains, $N_{i}$. We would like to find the distribution
over the thinned NNs which we denote $p(\mathcal{L_{\text{i}}},N_{i})$
or simply $p_{i}$. We can estimate the expected size $N$, as it
depends on the dropout value and the expected loss using the training
error. Thus, we can approximate the following two expectations 
\begin{equation}
\mathbb{E}[\mathcal{L}]=\sum_{i}p\left(\mathcal{L}_{i},N_{i}\right)\mathcal{L}_{i}=\mathcal{L}\label{eq:avrg_loss}
\end{equation}
\begin{equation}
\mathbb{E}[N]=\sum_{i}p\left(\mathcal{L}_{i},N_{i}\right)N_{i}=\mathcal{N}\label{eq:avrg_N}
\end{equation}

According to the principle of maximum entropy, \cite{jaynes1957information}
the distribution with maximal entropy best represents the current
state of knowledge (i.e., $-\sum p_{i}\log p_{i}$) when equations
\ref{eq:avrg_loss}, \ref{eq:avrg_N} and the normalization constraint
$\sum p_{i}=1$ are satisfied. To find the maximally unbiased probability
distribution, we define the following Lagrangian 

\begin{align*}
J[p] & =-\sum_{i}p_{i}\log p_{i}+\lambda_{0}\left(\sum_{i}p_{i}-1\right)+\beta\left(\sum_{i}p_{i}\mathcal{L}_{i}-\mathcal{L}\right)+\\
 & +\eta\left(\sum_{i}p_{i}N_{i}-\mathcal{N}\right)
\end{align*}

Setting the derivative $\frac{\partial J}{\partial p_{i}}$ to $0$,
we have that the probability of the $i$-th thinned NN is 
\begin{equation}
p_{i}=\frac{1}{Z}e^{-\left(\beta\mathcal{L}_{i}+\eta N_{i}\right)}\label{eq:boltz_GC}
\end{equation}
where $Z$ is the well-known partition function that normalizes the
probability measure $p_{i}$ 
\begin{equation}
Z=\sum_{i}e^{-\left(\beta\mathcal{L}_{i}+\eta N_{i}\right)}\label{eq:Z_GC}
\end{equation}
This is known as the Gibbs distribution in statistical mechanics,
where $\beta$ is known as the inverse temperature and $\eta$ is
the chemical potential. 

We can deduce the variance of the loss from $Z$,
\begin{equation}
\mathbb{V\text{ar}}[\mathcal{L}^{2}]=\frac{\partial^{2}\log Z}{\partial\beta^{2}}\label{eq:var_L}
\end{equation}

This variance is a measure of the uncertainty. Thus, we want to compute
the partition function. However, the partition function contains an
exponential number of terms and therefore is impossible to calculate
directly. In the next section, we present an algorithm that approximates
the value of this partition function using an importance sampling
technique; this technique gives a measure of the uncertainty with
respect to different cumulants. The partition function \ref{eq:Z_GC}
corresponds exactly to the well-known grand canonical partition function\cite{kardar2007statistical},
establishing the relationship between the two fields. While the partition
function \ref{eq:Z_GC} can generate moments to give a representation
of the uncertainty, obtaining this representation involves summing
over an exponential probability space. Sampling such a high dimensional
space is impractical; instead, we can use the partition function as
a scale. In the following sections, we present our algorithm, which
combines the probability measure \ref{eq:boltz_GC} with importance
sampling to obtain the variance in the loss $\mathcal{L}$ and number
of units $N$. This allows us to estimate the uncertainty using the
above moments.

\subsection{Loss Variance Monte Carlo Estimate}

It is impossible to calculate the cumulant analytically, but it is
possible to evaluate the cumulant numerically. We outline our Loss
Variance Monte Carlo Estimate (LoVME) algorithm to calculate the uncertainty.
We can write the estimator of $\mathbb{E}\left[Q\right]$ over $M$
samples for an observable $Q$ as $\left\langle Q\right\rangle ^{M}=\frac{\sum_{j=1}^{M}Q_{j}p_{j}}{\sum_{j=1}^{M}p_{j}}$
and see that as $M\rightarrow\infty$ this equation gives the exact
average of the estimator. As with previous methodologies using dropout,
one can choose to uniformly sample the $M$ states in the estimator.
Uniform sampling is equivalent to generating a network through a Bernoulli
trial much like dropout, this network will by the nature of this trial
take a long time to converge, since one can get stuck in a local minima
of this ``state space'' .

For more robust sampling of this estimator, we used the probability
measure \ref{eq:boltz_GC} as a non-uniform sampler. This approach
gives less weight to highly loss-thinned networks, and allows us to
avoid sampling only in a small region of thinned NN. To ensure that
we pick each state with its associated probability measure, we generate
a Markov chain Monte Carlo in an ergodic way, such that the rates
obey a detailed balance. Imposing the condition of ergodicity on our
Monte Carlo allows us to make some of the transitions in the Markov
chain zero, while requiring that there be at least one path of non-zero
transition probabilities between any two configurations. This forces
all states to be accessible. Imposing a second condition of detailed
balance, allows us to sample thinned networks with Metropolis-Hastings
algorithms. Requiring detailed balance ensures that after enough sampling
steps, we indeed sample with \ref{eq:boltz_GC}. Mathematically, we
can write down the transition dynamics between two thinned NNs as
\[
\sum_{v}p_{\mu}P\left(\mu\rightarrow v\right)=\sum_{v}p_{v}P\left(v\rightarrow\mu\right)
\]
here $P\left(\mu\rightarrow v\right)$ is the rate of transition.
Detailed balance ensures that all the thinned NN are accessible one
from another 
\[
p_{\mu}P\left(\mu\rightarrow v\right)=p_{v}P\left(v\rightarrow\mu\right)\ \forall\mu,v
\]
 This tells us that, on average the system should go from $\mu$ to
$v$ as much as it goes from $v$ to $\mu$. In fact, we can now use
the probability \ref{eq:boltz_GC} to write down the rates that will
generate the desired Markov chain.
\begin{equation}
\frac{P\left(\mu\rightarrow v\right)}{P\left(v\rightarrow\mu\right)}=\frac{p_{v}}{p_{\mu}}=e^{-\eta\left(N_{v}-N_{\mu}\right)}e^{-\beta\left(\mathcal{L}_{v}-\mathcal{L_{\mu}}\right)}\label{eq:DB}
\end{equation}
The constraint of \ref{eq:DB} still leaves us a good deal of freedom
over how to choose the transition probabilities, since there are many
ways to satisfy the equality. Rewriting the detailed balance condition
\[
\frac{P\left(\mu\rightarrow v\right)}{P\left(v\rightarrow\mu\right)}=\frac{g\left(\mu\rightarrow v\right)}{g\left(v\rightarrow\mu\right)}\frac{A\left(\mu\rightarrow v\right)}{A\left(v\rightarrow\mu\right)}
\]
 we can introduce the acceptance function $A$ for the Monte Carlo
move and selection probabilities $g$. The following lets us look
at the reversible selection probabilities of $\frac{1}{N}$ and 
\[
A\left(v\rightarrow\mu\right)=A_{0}e^{-\frac{1}{2}\eta\left(N_{\mu}-N_{v}\right)}e^{-\beta\left(\mathcal{L}^{\mu}-\mathcal{L}^{v}\right)}
\]
 to obtain the acceptance rule in the form 
\begin{equation}
A\left(v\rightarrow\mu\right)=A_{0}e^{-\frac{1}{2}\eta\left(N_{\mu}-N_{v}\right)}e^{-\beta\left(\mathcal{L}^{\mu}-\mathcal{L}^{v}\right)}\label{eq:MC_acc}
\end{equation}
By building on \ref{eq:MC_acc} and using Metropolis rates we can
write the full acceptance rule for our simulations 
\begin{equation}
A\left(v\rightarrow\mu\right)=\begin{cases}
e^{-\beta\left(\left(\mathcal{L}^{\mu}-\mathcal{L}^{v}\right)+\frac{\eta}{\beta}\left(N_{\mu}-N_{v}\right)\right)} & \text{if }\mathcal{L}^{\mu}-\mathcal{L}^{v}+\frac{\eta}{\beta}\left(N_{\mu}-N_{v}\right)>0\\
1 & \text{otherwise }
\end{cases}\label{eq:GCMC_acc}
\end{equation}
Using sequential dynamical Monte Carlo moves with the acceptance rate
\ref{eq:GCMC_acc} we can generate the ensemble of different thinned
NNs. In Algorithm \ref{alg:LOVEME}, we give the flow and the details
of the LoVME algorithm implemented with the above rates. 

\begin{algorithm}[h]
\caption{Loss Variance Monte Carlo Estimate ((LoVME)}\label{euclid} \begin{algorithmic}[1] \STATE \textbf{Input:} Oracle access to loss $\mathcal{L}_i$ for each thinned NN $i$  \STATE \textbf{Output:} Estimate $\mathbb{V}ar[\mathcal{L}_i]$ \STATE \textbf{Parameter:} $T$ - number of transitions the algorithm performs  \STATE \textbf{Initialization:}  \STATE $1. L$: array of size $T$ of losses encountered throughout the run of the algorithm \STATE $2. N_0, \mathcal{L}_0$: loss and number of neurons in the fully connected trained NN \STATE $3. N_\mu = N_0$: first NN in HM algorithm  \STATE $4. t=1$ \WHILE{$t \leq T$} \STATE $N \in_R[N_0]$, random from the discrete uniform distribution  \STATE $v :=$ random thinned NN with $N$ neurons  \STATE $\theta\sim U(0,1)$, random from the continuous  uniform distribution  \IF {$\mathcal{L}_\mu - \mathcal{L}_v + \frac{\eta}{\beta}(N_\mu - N_v) < \theta$} \STATE $L[t] = \mathcal{L}_v$ \STATE $\mu := v$ \STATE $t := t + 1$ \ENDIF \ENDWHILE  \RETURN variance of array $L$ \end{algorithmic}

\caption{\label{alg:LOVEME}Loss Variance Monte Carlo Estimate}
\end{algorithm}

\section{Experiments}

We now demonstrate the ability of the LoVMe algorithm to predict
model uncertainty with respect to several different metrics. For in-depth
insight into the structure of the uncertainty measure in our ensemble,
we qualitatively evaluated the higher order cumulates for the loss
distribution. In addition to evaluating several different metrics
for the prediction, we also evaluated a series of higher order cumulates
and their effects on the uncertainty. To estimate the uncertainty,
we trained a LeNET \cite{lecun1998gradient} network in PyTorch \cite{paszke2017automatic},
using dropout with probability $p=0.5$ \footnote{Code to be available for both the network and the LoVMe}.
We used a cross-entropy softmax loss function in both the training
and testing phases. For the testing stage and to calculate the Monte
Carlo moves, we evaluated the loss function against a randomly generated
label set, for both the uncertainty measure against $N$ and $\mathcal{L}$.

\subsection{Data }

To demonstrate the effectiveness of the LoVEMe algorithm \ref{alg:LOVEME},
we applied it to three different datasets: the widely adopted MNIST
\cite{lecun1998mnist}, Fashion-MNIST \cite{xiao2017/online}, and
CIFAR-10 \cite{krizhevsky2009learning} . For both the MNIST and Fashion-MNIST
we also tested against a partially perturbed subset, where we introduced
both random rotation and noise to randomly chosen $10\%$ of the images
in the test set. For each dataset, we generated an ensemble of thinned
NNs using the LoVMe algorithm; each of these ensembles typically converged
within 300 Monte Carlo steps. 

\subsection{Obtaining a Ground Truth }

\begin{figure}
\begin{centering}
\subfloat[\label{fig:higher_moments}The Monte Carlo loss distribution for two
data sets studied. \protect \\
Each distribution is color coded to their image.]{\begin{centering}
\includegraphics[width=0.4\paperwidth]{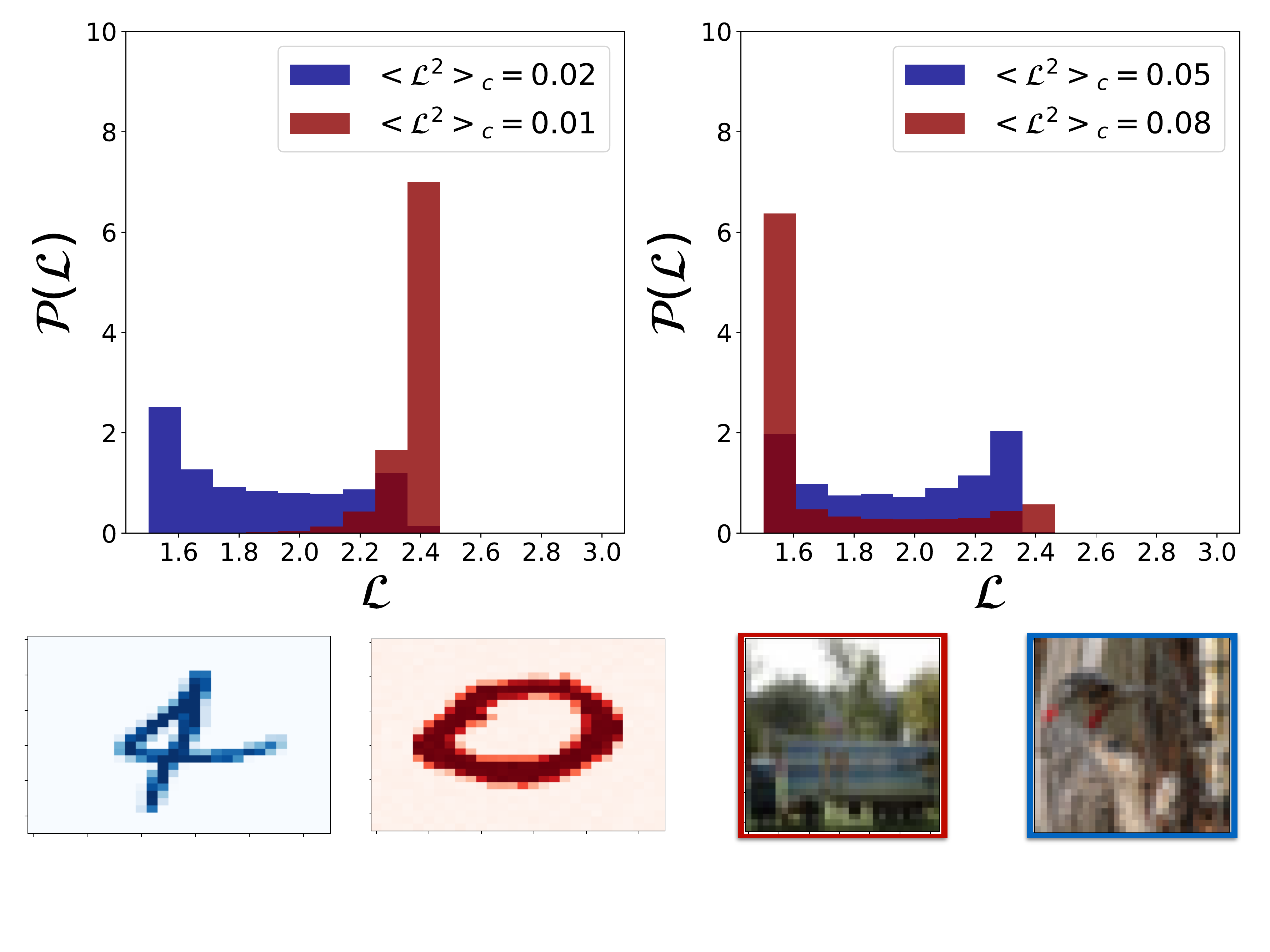}
\par\end{centering}
}$ $$ $$ $$ $\subfloat[\label{fig:The-loss-variance_corr}Our ``ground truth'' uncertainty,
the loss variance and average probability correlation function ]{\includegraphics[width=0.2\paperwidth]{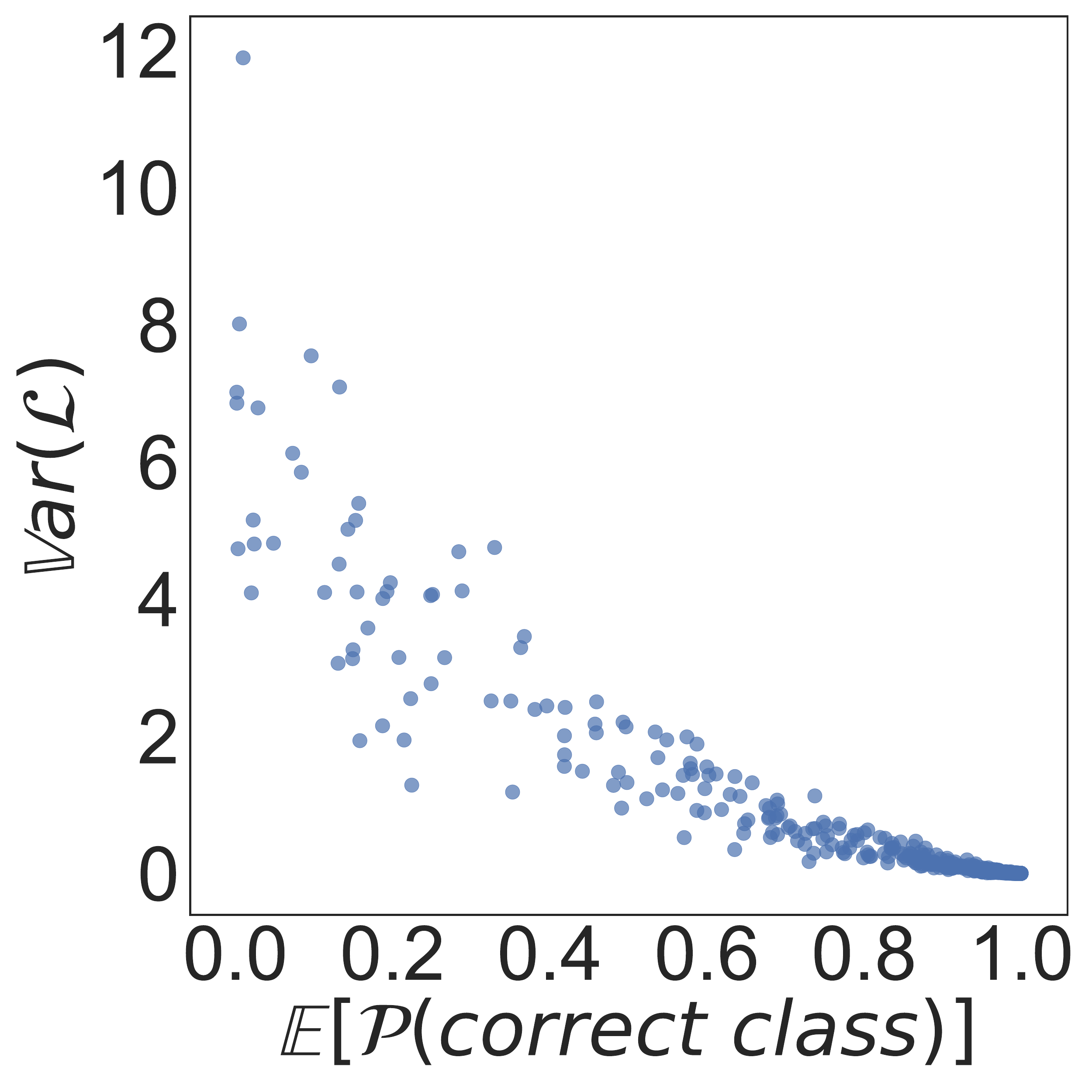}}
\par\end{centering}
\caption{Monte Carlo results for uncertainty }
\end{figure}

Before describing our uncertainty measure in detail, it's worthwhile
to obtain an understanding of the intrinsic or ground truth uncertainty
that a model possess. To gauge the basic uncertainty measure, we constructed
a ``ground truth'' measure of the uncertentity. This was done by
training an ensemble of $4000$ LeNET networks with different initializations
of the weights. For each ensemble we performed a classification of
test portion in the dataset, and took the probability outcome from
the softmax cross-entropy for correct class. 

Fig. \ref{fig:The-loss-variance_corr} shows the scatter plot between
the variance in the loss, the uncertainty $\mathbb{{V}}ar[\mathcal{L}]$
and the average over the different probabilities of the correct label
$\mathbb{E}[P\left(\text{correct label}\right)$. It is interesting
to note that the point at which the probability of the model returning
the correct label equals 1 corresponds to a transition point between
the uncertainty and the probability being correlated and uncorrelated.
Furthermore, there is a region where for a high uncertainty we have
an ambiguous or almost random probability of the model returning the
correct prediction. 

\subsection{Estimating the Uncertainty using LoVME}

We next estimate the uncertainty using the variance $\mathbb{{V}}ar[N]$
and $\mathbb{{V}}ar[\mathcal{L}]$. In Fig. \ref{fig:higher_moments}
we show two distributions of $\mathcal{L}$ and their respective test
examples for the CIFAR10 and MNIST dataset. The MNIST data set quantitatively
showed lower values of variance. This is to be expected as the network
learns and converges within a few epochs to $99.7\%$ accuracy. Nonetheless,
by plotting the distribution and calculating its variance, we can
see the variance and uncertainty of each sample. Each color in Fig.
\ref{fig:higher_moments} coincides with its distribution function,
and the figure legend shows the variance values. For the CIFAR data
set, there is a more distinct difference in the uncertainty values
and in the spread of the distribution. It is also informative to look
at the structure of a few chosen examples and their distributions,
as shown in Fig. \ref{fig:The-loss-variance_corr} shows the structure
of a few chosen examples and their distributions. Surprisingly, most
distributions either look bimodal or non-symmetric, which suggests
that while the variance measures the uncertainty, higher moments might
also come into play while considering a sample's uncertainty. Our
LoVME method demonstrates this important fact by evaluating the distribution
explicitly, as well as by looking at higher moments. 

\subsection{Uncertainty in a Decision Metric}

Estimating the uncertainty during the test stage also enables us to
incorporate uncertainty into the decision metric. To show the effect
of uncertainty on the prediction, we show the loss variance scores
through both Receiver Operating Characteristic (ROC) and Area Under
the Curve (AUC) plots. 

\begin{figure}
\begin{centering}
\includegraphics[width=0.25\paperwidth]{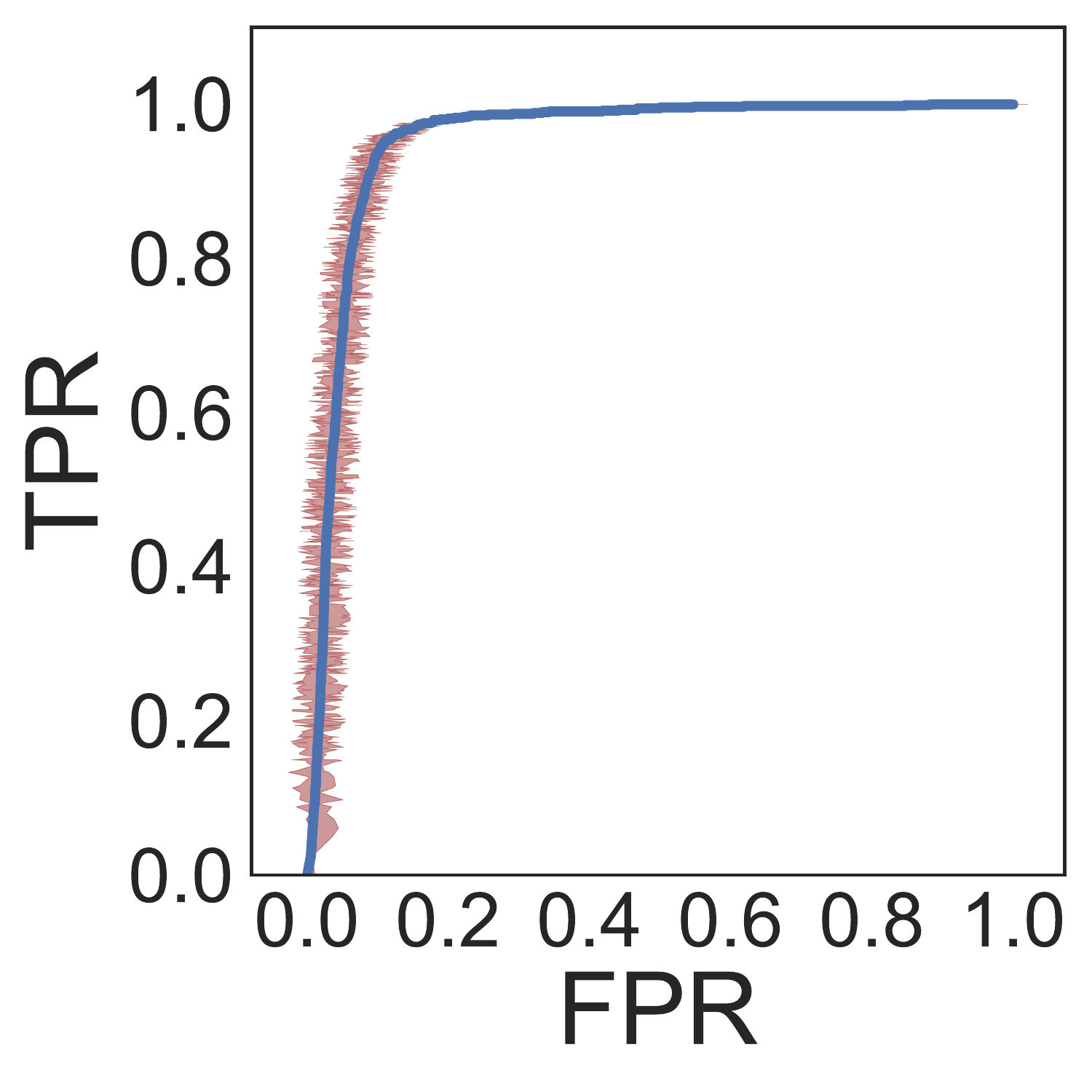}\includegraphics[width=0.25\paperwidth]{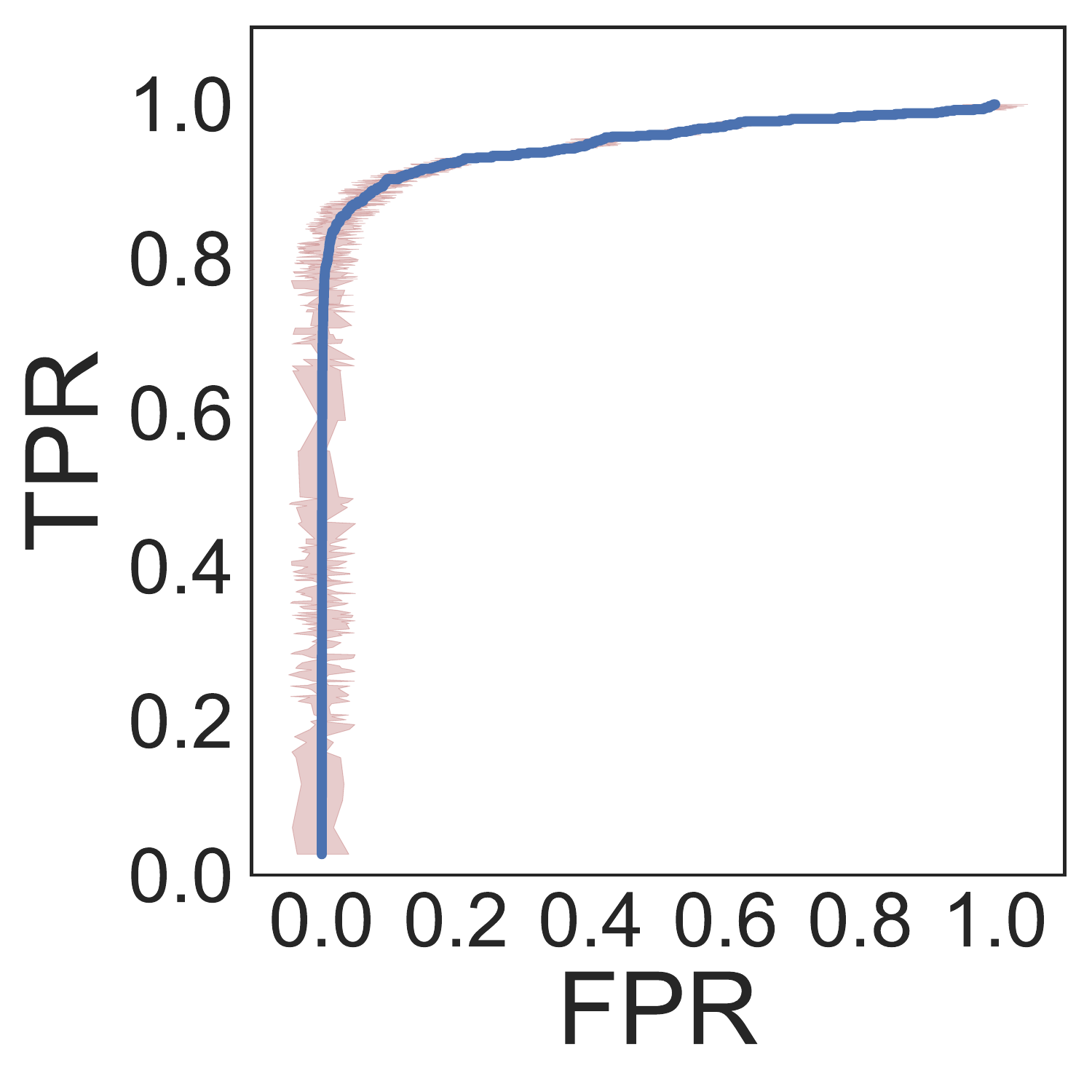} 
\par\end{centering}
\caption{\label{fig:ROC_fuzzy}ROC for two classes, one of each dataset, with
our ``confidence intervals'' measures}
\end{figure}

\begin{figure}
\begin{centering}
\includegraphics[width=0.25\paperwidth]{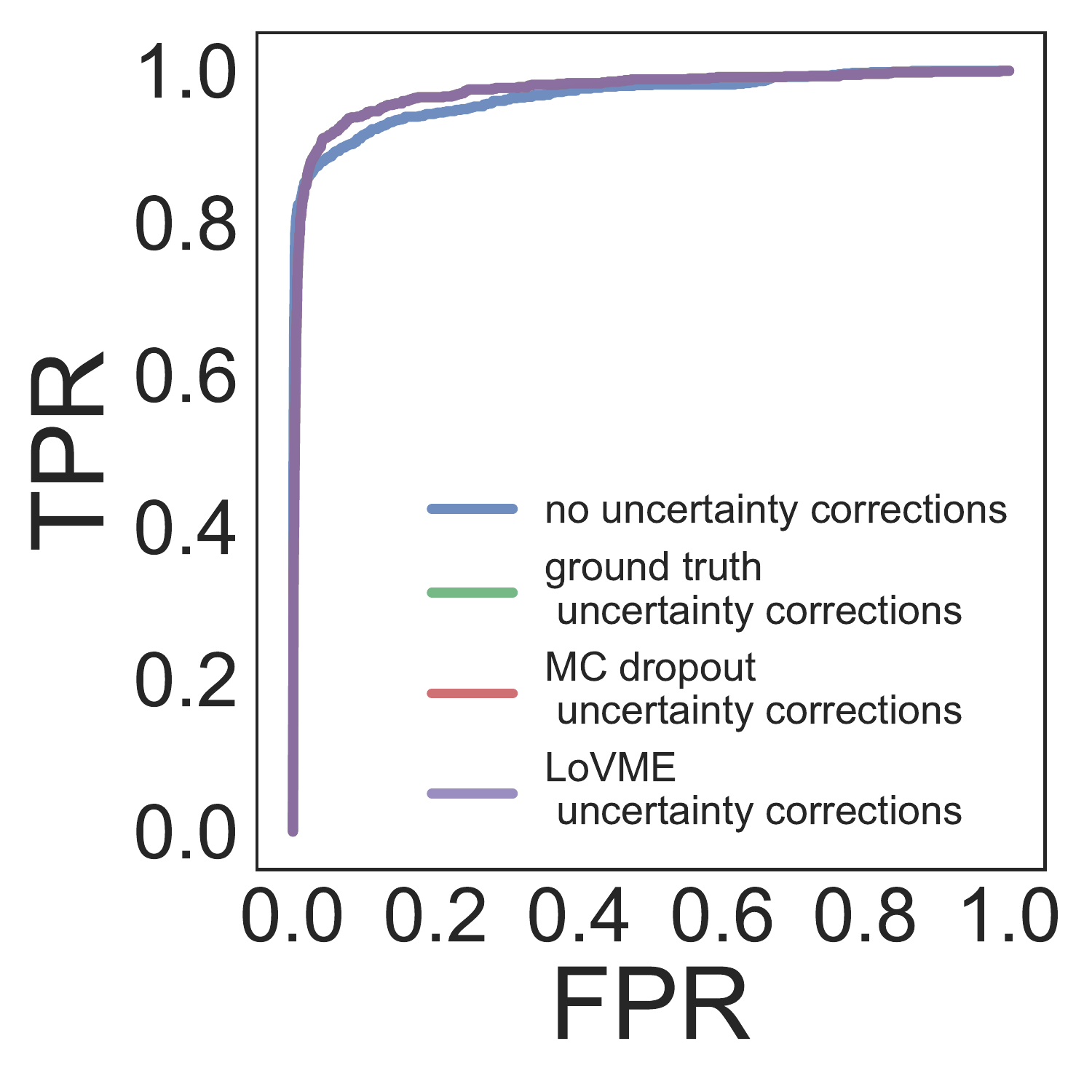}\includegraphics[width=0.25\paperwidth]{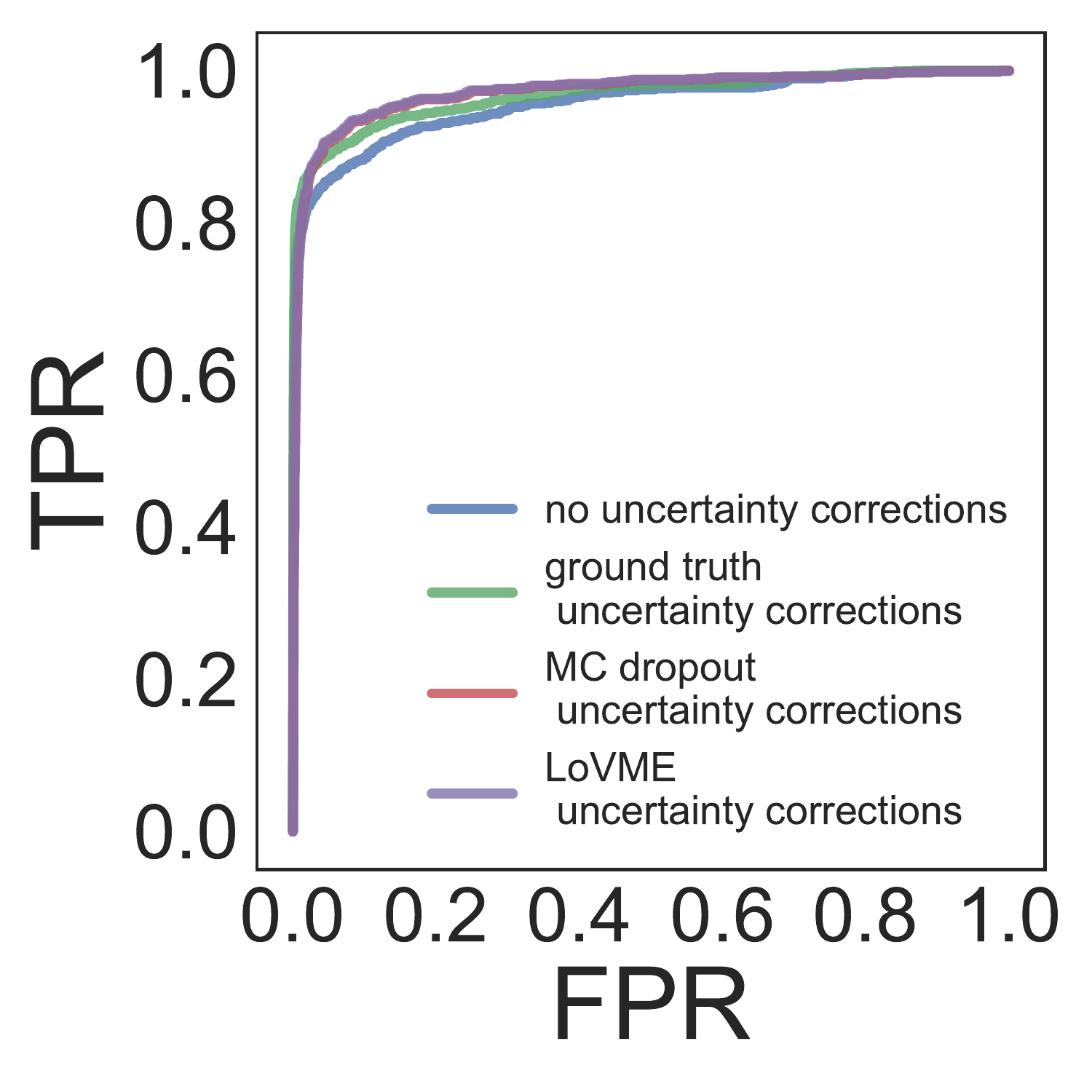}
\par\end{centering}
\caption{\label{fig:ROC_all}ROC curves with improvements based on uncertainty
measures }
\end{figure}

For a probability score $P$ of a sample, we use its associated uncertainty
as a binary confidence interval. Looking at $P\pm\mathbb{V}ar\left[\mathcal{L}^{2}\right]$
could either shift the sample into a TPR or a FPR, allowing us to
use a type of confidence interval on top of the ROC, as shown in Fig.
\ref{fig:ROC_fuzzy}.Generating the confidence interval of the score
allows us to set a certain threshold, above which a sample is too
ambiguous to be classified as either a false positive or true negative.
Thus, it can be eliminated from the ROC decision process. In Fig.
\ref{fig:ROC_all} we show that removing high uncertainty examples
increases the AUC when using our LoVME method, Monte Carlo dropout,
and when calculating the naïve ground truth. Table \ref{tab:AUC-and-ROC}
shows the different values of the AUC for each of the three methods. 

\begin{table}[H]
\begin{centering}
\begin{tabular}{|c|>{\centering}p{3cm}|>{\centering}p{3cm}|>{\centering}p{3cm}|>{\centering}p{3cm}|}
\hline 
 & \multicolumn{1}{>{\centering}p{3cm}||}{no uncertainty corrections} & \multicolumn{1}{>{\centering}p{3cm}||}{ground truth uncertainty corrections} & \multicolumn{1}{>{\centering}p{3cm}||}{MC dropout uncertainty corrections} & LoVME $ $ uncertainty corrections\tabularnewline
\hline 
\hline 
MNIST AUC & 0.967 & 0.974 & 0.974 & 0.974\tabularnewline
\hline 
CIFAR10 AUC & 0.956 & 0.967 & 0.972 & 0.973\tabularnewline
\hline 
\end{tabular}
\par\end{centering}
\caption{\label{tab:AUC-and-ROC}AUC values for the set of experiments }
\end{table}

\bibliographystyle{plain}
\bibliography{bibfile}

\begin{thebibliography}{10}

\bibitem{baity2018comparing}
M~Baity-Jesi, L~Sagun, M~Geiger, S~Spigler, G~Ben Arous, C~Cammarota, Y~LeCun,
  M~Wyart, and G~Biroli.
\newblock Comparing dynamics: Deep neural networks versus glassy systems.
\newblock {\em arXiv preprint arXiv:1803.06969}, 2018.

\bibitem{bellazzi2008predictive}
Riccardo Bellazzi and Blaz Zupan.
\newblock Predictive data mining in clinical medicine: current issues and
  guidelines.
\newblock {\em International journal of medical informatics}, 77(2):81--97,
  2008.

\bibitem{cho2009kernel}
Youngmin Cho and Lawrence~K Saul.
\newblock Kernel methods for deep learning.
\newblock In {\em Advances in neural information processing systems}, pages
  342--350, 2009.

\bibitem{choromanska2015loss}
Anna Choromanska, Mikael Henaff, Michael Mathieu, G{\'e}rard~Ben Arous, and
  Yann LeCun.
\newblock The loss surfaces of multilayer networks.
\newblock In {\em Artificial Intelligence and Statistics}, pages 192--204,
  2015.

\bibitem{fox2012tutorial}
Charles~W Fox and Stephen~J Roberts.
\newblock A tutorial on variational bayesian inference.
\newblock {\em Artificial intelligence review}, 38(2):85--95, 2012.

\bibitem{gal2016uncertainty}
Yarin Gal.
\newblock Uncertainty in deep learning.
\newblock {\em University of Cambridge}, 2016.

\bibitem{gal2016dropout}
Yarin Gal and Zoubin Ghahramani.
\newblock Dropout as a bayesian approximation: Representing model uncertainty
  in deep learning.
\newblock In {\em international conference on machine learning}, pages
  1050--1059, 2016.

\bibitem{hanley1982meaning}
James~A Hanley and Barbara~J McNeil.
\newblock The meaning and use of the area under a receiver operating
  characteristic (roc) curve.
\newblock {\em Radiology}, 143(1):29--36, 1982.

\bibitem{jaynes1957information}
Edwin~T Jaynes.
\newblock Information theory and statistical mechanics.
\newblock {\em Physical review}, 106(4):620, 1957.

\bibitem{kardar2007statistical}
Mehran Kardar.
\newblock {\em Statistical physics of fields}.
\newblock Cambridge University Press, 2007.

\bibitem{kendall2015bayesian}
Alex Kendall, Vijay Badrinarayanan, and Roberto Cipolla.
\newblock Bayesian segnet: Model uncertainty in deep convolutional
  encoder-decoder architectures for scene understanding.
\newblock {\em arXiv preprint arXiv:1511.02680}, 2015.

\bibitem{kendall2017uncertainties}
Alex Kendall and Yarin Gal.
\newblock What uncertainties do we need in bayesian deep learning for computer
  vision?
\newblock In {\em Advances in Neural Information Processing Systems}, pages
  5580--5590, 2017.

\bibitem{krizhevsky2009learning}
Alex Krizhevsky and Geoffrey Hinton.
\newblock Learning multiple layers of features from tiny images.
\newblock 2009.

\bibitem{lecun1998mnist}
Yann LeCun.
\newblock The mnist database of handwritten digits.
\newblock {\em http://yann. lecun. com/exdb/mnist/}, 1998.

\bibitem{lecun2015deep}
Yann LeCun, Yoshua Bengio, and Geoffrey Hinton.
\newblock Deep learning.
\newblock {\em nature}, 521(7553):436, 2015.

\bibitem{lecun1990handwritten}
Yann LeCun, Bernhard~E Boser, John~S Denker, Donnie Henderson, Richard~E
  Howard, Wayne~E Hubbard, and Lawrence~D Jackel.
\newblock Handwritten digit recognition with a back-propagation network.
\newblock In {\em Advances in neural information processing systems}, pages
  396--404, 1990.

\bibitem{lecun1998gradient}
Yann LeCun, L{\'e}on Bottou, Yoshua Bengio, and Patrick Haffner.
\newblock Gradient-based learning applied to document recognition.
\newblock {\em Proceedings of the IEEE}, 86(11):2278--2324, 1998.

\bibitem{lisboa2006use}
Paulo~J Lisboa and Azzam~FG Taktak.
\newblock The use of artificial neural networks in decision support in cancer:
  a systematic review.
\newblock {\em Neural networks}, 19(4):408--415, 2006.

\bibitem{mackay1996near}
David~JC MacKay and Radford~M Neal.
\newblock Near shannon limit performance of low density parity check codes.
\newblock {\em Electronics letters}, 32(18):1645, 1996.

\bibitem{paszke2017automatic}
Adam Paszke, Sam Gross, Soumith Chintala, Gregory Chanan, Edward Yang, Zachary
  DeVito, Zeming Lin, Alban Desmaison, Luca Antiga, and Adam Lerer.
\newblock Automatic differentiation in pytorch.
\newblock 2017.

\bibitem{schoenholz2017correspondence}
Samuel~S Schoenholz, Jeffrey Pennington, and Jascha Sohl-Dickstein.
\newblock A correspondence between random neural networks and statistical field
  theory.
\newblock {\em arXiv preprint arXiv:1710.06570}, 2017.

\bibitem{srivastava2014dropout}
Nitish Srivastava, Geoffrey Hinton, Alex Krizhevsky, Ilya Sutskever, and Ruslan
  Salakhutdinov.
\newblock Dropout: A simple way to prevent neural networks from overfitting.
\newblock {\em The Journal of Machine Learning Research}, 15(1):1929--1958,
  2014.

\bibitem{xiao2017/online}
Han Xiao, Kashif Rasul, and Roland Vollgraf.
\newblock Fashion-mnist: a novel image dataset for benchmarking machine
  learning algorithms, 2017.

\end{thebibliography}
 
\end{document}